\documentclass[9pt,twocolumn,twoside,lineno]{gdm/gdm-new}

\templatetype{gdm/gdmresearcharticle}
\usepackage{graphicx}
\usepackage{multirow}
\usepackage{array}
\usepackage{subcaption}
\usepackage{amssymb} %
\usepackage{pifont} %
\usepackage{makecell}
\usepackage{algorithm}
\usepackage{algpseudocode}
\usepackage{amsmath}
\usepackage{cleveref}
\usepackage{multicol}
\usepackage{float}
\usepackage{lineno}
\newif\ifarxiv
\arxivtrue 
\newcommand{\extlink}[2]{\ifarxiv\href{#1}{#2}\fi}
\newcommand{\removeinside}[1]{\ifarxiv #1\fi}
\newcommand{\ourtitle}{Do generative video models\\understand physical principles?}
\newcommand{\oursection}[1]{\section*{#1}}
\newcommand{\ourintrosection}[1]{}
\newcommand{\oursubsection}[1]{\subsection*{#1}}
\newcommand{\ourabstract}[1]{#1}
\title{\titlefont{\ourtitle}}

\let\DefaultHeadRule\headrule
\let\DefaultFootRule\footrule
\renewcommand{\headrule}{\color{gray}\DefaultHeadRule}
\renewcommand{\footrule}{\textcolor{gray}{\DefaultFootRule}}
\usepackage{fancyhdr}
\fancypagestyle{firststyle}{
	\fancyhead[L]{
			\includegraphics[width=90pt]{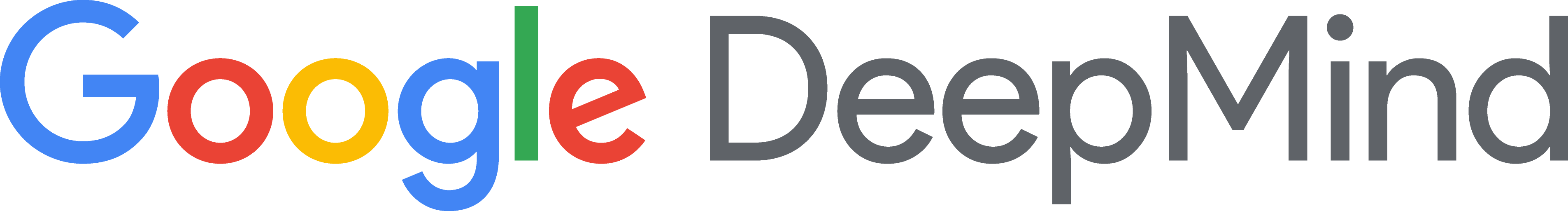}
}
\fancyhead[R]{\footerfont\itshape\today
}
\fancyhead[C]{}
}

\author[]{\large Saman Motamed\textsuperscript{a,1}, Laura Culp\textsuperscript{b}, Kevin Swersky\textsuperscript{b}, Priyank Jaini\textsuperscript{b,$\dagger$}, and Robert Geirhos\textsuperscript{b,$\dagger$}\\\normalsize\textsuperscript{a}INSAIT, Sofia University; work done while at Google DeepMind; \textsuperscript{b}Google DeepMind; \textsuperscript{$\dagger$}Joint last authors.}

\leadauthor{Motamed}

\begin{document}

\date{}
\maketitle

\thispagestyle{firststyle}

{\absfont{\noindent\ourabstract{
AI video generation is undergoing a revolution, with quality and realism advancing rapidly. These advances have led to a passionate scientific debate: Do video models learn ``world models'' that discover laws of physics---or, alternatively, are they merely sophisticated pixel predictors that achieve visual realism without understanding the physical principles of reality? We address this question by developing Physics-IQ, a comprehensive benchmark dataset that can only be solved by acquiring a deep understanding of various physical principles, like fluid dynamics, optics, solid mechanics, magnetism and thermodynamics. We find that across a range of current models (Sora, Runway, Pika, Lumiere, Stable Video Diffusion, and VideoPoet), physical understanding is severely limited, and unrelated to visual realism. At the same time, some test cases can already be successfully solved. This indicates that acquiring certain physical principles from observation alone may be possible, but significant challenges remain. While we expect rapid advances ahead, our work demonstrates that visual realism does not imply physical understanding. \removeinside{Our project page is at \href{https://physics-iq.github.io/}{Physics-IQ-website}; code at \href{https://github.com/google-deepmind/physics-IQ-benchmark}{Physics-IQ-benchmark}.}}
}}
\begin{figure*}
    \centering
    \includegraphics[width=\textwidth]{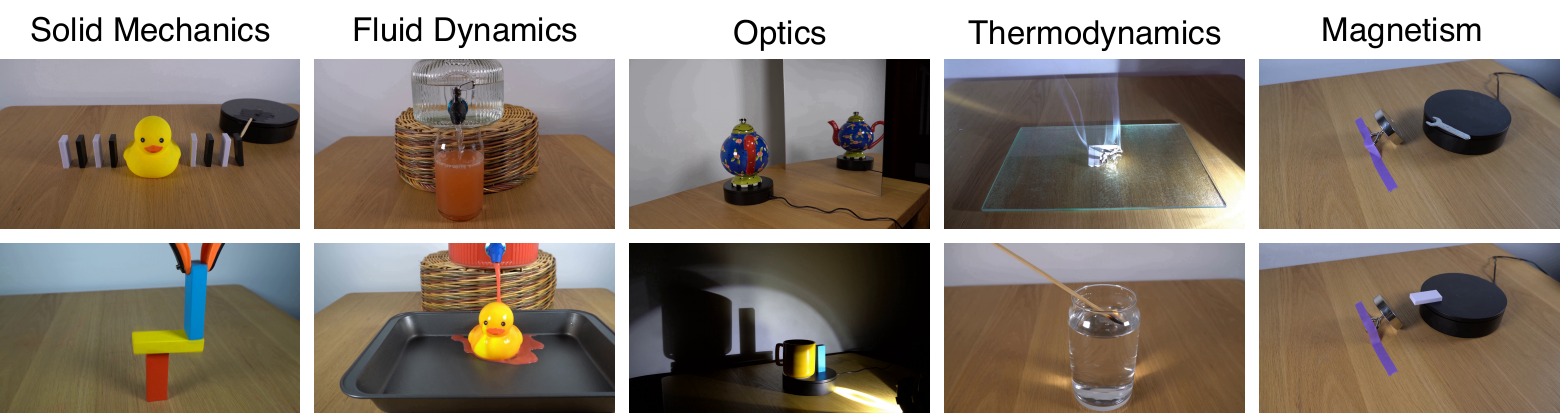}
    \caption{Sample scenarios from the Physics-IQ dataset for testing physical understanding in generative video models. Models are shown the beginning of a video (single frame for image2video models; 3 seconds for video2video models)  and need to predict how the video continues over the next 5 seconds, which requires understanding different physical properties: Solid Mechanics, Fluid Dynamics, Optics, Thermodynamics, and Magnetism. \removeinside{See \extlink{https://physics-iq.github.io/figures/fig1.html}{here} for an animated version of this figure.}}
    \label{fig:our_dataset_qualitative}
\end{figure*}

\ourintrosection{Introduction}
Can a machine truly understand the world without interacting with it? This question lies at the heart of the ongoing debate surrounding the capabilities of AI video generation models. While the generation of realistic videos has, for a long time, been considered one of the major unsolved challenges within deep learning, this recently changed. Within a relatively short period of time, the field has seen the development of impressive video generation models \cite{sora_openai, veo2_deepmind, meta_movie_gen}, capturing the imagination of the public and researchers alike. A major milestone towards general-purpose artificial intelligence is to build machines that understand the world, and if you cannot understand what you cannot create (as Feynman would say), then the ability of those models to create visually realistic scenes is an essential step towards that capability. However, the degree to which successful \emph{generation} signals successful \emph{understanding} is the subject of a passionate debate. Is it possible to understand the world without ever interacting with it? Phrased differently, do generative video models learn the physical principles that underpin reality from ``watching'' videos?

Proponents argue that the way the models are trained---predicting how videos continue, a.k.a.\ next frame prediction---is a task that forces models to understand physical principles. According to this line of argument, it is impossible to predict the next frame of a sequence if the model has no understanding of how objects move (trajectories), that things fall down instead of up (gravity), and how pouring juice into a glass of water changes its color (fluid dynamics). As an analogy, large language models are trained in a similar fashion to predict the next tokens (characters or words) in a text; a task formulation that is equally simple but has proven sufficient to enable impressive capabilities and text understanding. Moreover, predicting the future is a core principle of biological perception, too: The brain constantly generates predictions about incoming sensory input, enabling energy-efficient processing of information \citep{barlow1961possible} and building a mental model of the world as postulated by von Helmholtz \citep{von1867handbuch} and later the predictive coding hypothesis \citep{friston2005theory}. In short, successful prediction signals successful understanding.

On the other hand, there are also important arguments contra understanding through observation. According to the causality rationale, ``watching'' videos (or to be more precise, training models to predict how videos continue) is a passive process, with models unable to interact with the world. This lack of interaction means that a model cannot observe the causal effects of an intervention (as, for instance, children are able to when playing with toys). 

Therefore, a model is faced with the nearly impossible task of distinguishing correlation from causation if it is to succeed in understanding physical principles.

Furthermore, video models that are touted as ``a promising path towards building general purpose simulators of the physical world'' \citep{sora_openai} arguably experience a different world to begin with: the digital world as opposed to the real world that an embodied system (like a robot, or virtually all living beings) experience. As a consequence, skeptics argue that visual realism by no means signals true understanding: All it takes to produce realistic videos is to reproduce common patterns from the model's vast sea of training data---shortcuts without understanding \citep{geirhos2020shortcut,kang2024far}.

In light of these two diametrically opposed perspectives, how can we tell whether generative video models indeed learn physical principles? To address this question in a quantifiable, tractable way, we created a challenging testbed for physical understanding in video models: the ``Physics-IQ'' benchmark. The core idea is to enlist video models to do what they do best: predict the continuation of a video. In order to test understanding, we designed a range of diverse scenarios where predicting the continuation requires a deep understanding of physical principles, going beyond pattern reproduction and testing out-of-distribution generalization. For instance, models are asked to predict how a domino chain falls---normally, vs.\ when a rubber duck is placed in the middle of the chain; or how pillows react when a kettlebell vs.\ a piece of paper is dropped onto the pillow. The diverse set of scenarios encompass solid mechanics, fluid dynamics, optics, thermodynamics and magnetism, totalling 396 high-quality videos filmed from three different perspectives in a controlled environment. Samples are shown in \Cref{fig:our_dataset_qualitative}. We then compare the model's prediction to the ground truth continuation using a set of metrics that capture different desiderata, and analyze a range of current models: Sora~\cite{sora_openai}, Runway Gen 3~\cite{runway2024}, Pika 1.0~\cite{pikalabs2024}, Lumiere~\cite{bartal2024lumiere}, Stable Video Diffusion~\cite{blattmann2023stable}, and VideoPoet~\cite{videopoet}.
\begin{figure*}[ht]
    \centering
    \includegraphics[width=\textwidth]{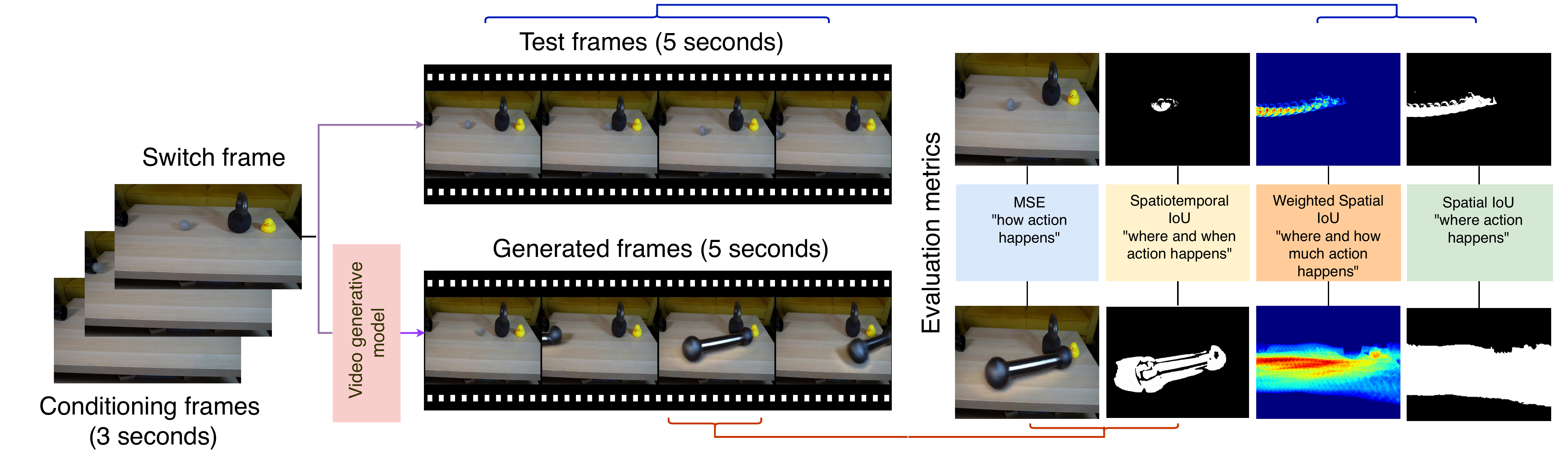}
    \caption{Overview of the Physics-IQ  evaluation protocol. A video generative model produces a 5 second continuation of the conditioning frame(s), optionally including a textual description of the conditioning frames for models that accept text input. They are compared against the ground truth test frames using four metrics that quantify different properties of physical understanding. The metrics are defined and explained in the methods section. \removeinside{Code to run the evaluation is available at \extlink{https://github.com/google-deepmind/physics-IQ-benchmark}{Physics-IQ-benchmark}.}}
    \label{fig:method}
\end{figure*}

\oursection{Physics-IQ benchmark}

\oursubsection{Dataset}
\label{subsec:dataset}
Our goal is to develop a dataset that tests the physical understanding capabilities of video generative models on different physical laws like solid mechanics, fluid dynamics, optics, thermodynamics, and magnetism. We therefore created the Physics-IQ dataset which consists of 396 videos each 8 seconds long covering 66 different physical scenarios. Each scenario in our dataset focuses on a specific physical law and aims to test a video generative model's understanding of physical events. These events include examples like collisions, object continuity, occlusion, object permanence, fluid dynamics, chain reactions, trajectories under the influence of forces (e.g., gravity), material properties and reactions, as well as lights, shadows, reflections, and magnetism.

\begin{figure}[ht]
    \centering
    \includegraphics[width=\columnwidth]{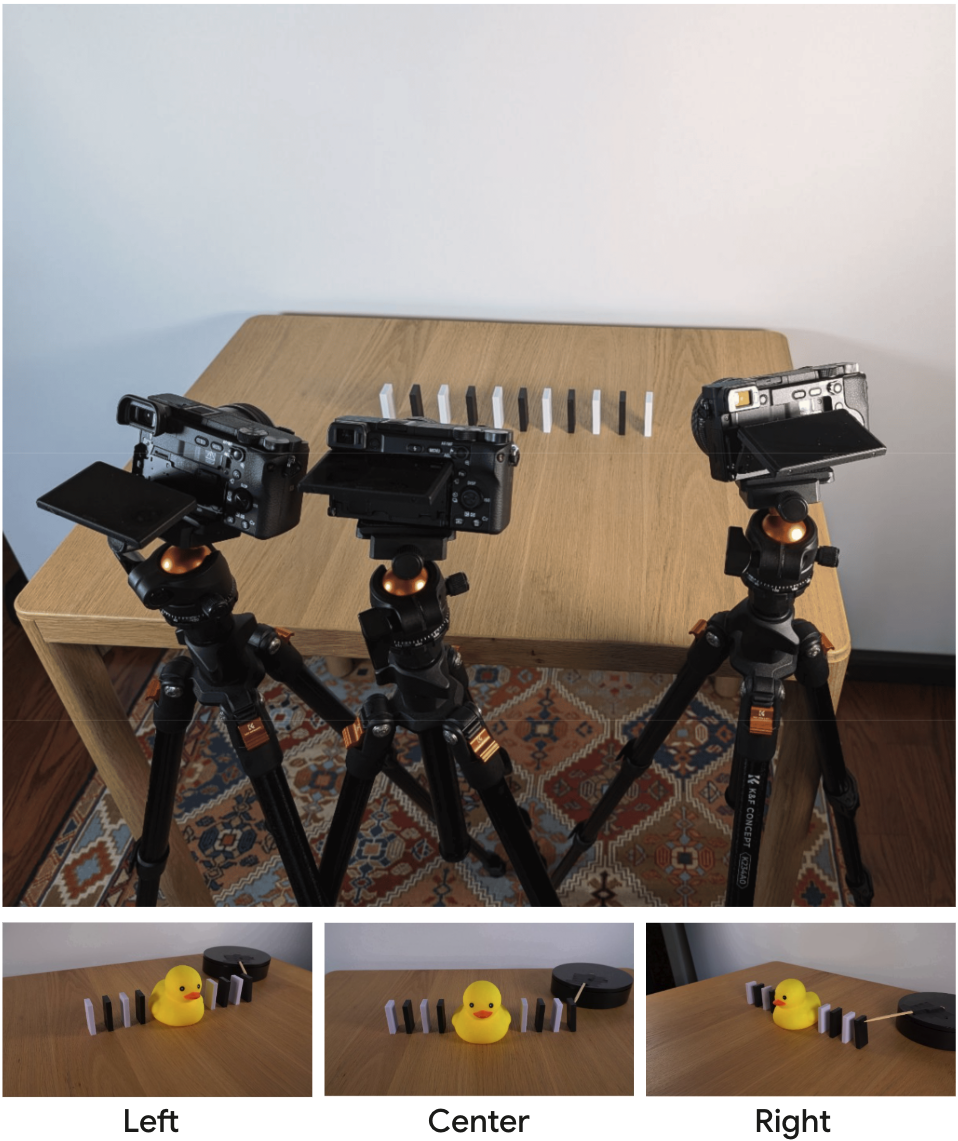}
    \caption{Illustration of recording setup (top) and perspectives (bottom).}
    \label{fig:setup}
\end{figure}

Each scenario was filmed at 30 frames per second (FPS) with a resolution of \(3840 \times 2160\) (16:9 aspect ratio) from three different perspectives: $\mathsf{left}$, $\mathsf{center}$, and $\mathsf{right}$ using high-quality Sony Alpha a6400 cameras equipped with 16-50mm lenses. Each scenario was shot twice (\textit{take 1} and \textit{take 2}) under identical conditions to capture the inherent variability of real-world physical interactions. These variations are expected in real-world due to factors like chaotic motion, subtle changes in friction, and variations in force trajectory. In this paper, we refer to the differences observed between these two recordings of the same scenario as \emph{physical variance}. This results in a total of 396 videos (66 scenarios $\times$ 3 perspectives $\times$ 2 takes). All our videos are shot from a static camera perspective without camera motion. The setup for filming the videos is illustrated in \cref{fig:setup}. \removeinside{The full dataset and code for evaluating model predictions is open-sourced here:\\ \url{https://github.com/google-deepmind/physics-IQ-benchmark}}

\oursubsection{Evaluation protocol}
\label{subsec:protocol}
Physical understanding can be measured in different ways. One of the most stringent tests is whether a model can predict how a challenging, unusual video continues---such as a chain of dominoes with a rubber duck in the middle interrupting the chain. Out-of-distribution scenarios like these test true understanding, since they cannot be solved by reproducing patterns seen or memorized from the training data \cite[e.g.][]{geirhos2018generalisation,hendrycks2018benchmarking,geirhos2020shortcut,srivastava2022beyond}. We therefore test physical understanding of video generative models by taking a full video of 8 seconds in which a physically interesting event occurs, splitting the video into a 3-second conditioning video and a 5-second test video which acts as ground truth. The model is then given the conditioning signal: either the 3-second video for video2video models (named \emph{multiframe models} in figures), or the last frame of this---called the \emph{switch frame}---in the case of image2video models (named \emph{i2v models} in figures). Since video models are trained precisely to generate the next frames given the previous frame(s) as conditioning signal, our evaluation protocol matches the paradigm these models were trained for. The switch frame is carefully selected manually for each scenario such that enough information about the physical event and objects in the scenario is provided, while at the same time making sure that successfully predicting the continuation requires some understanding of physics (e.g., in the scenario involving the chain reaction when a domino falls, the switch frame corresponds to the moment when the first domino is tipped but has not yet contacted the second domino). We provide video models that support multi-frame conditioning with as many conditioning frames (up to a maximum of 3 seconds) as they can accommodate. Some video models (e.g., Runway Gen 3, Pika 1.0, and Sora) generate subsequent frames based on a single image. For these models, we provide just the switch frame as the conditioning signal. \Cref{fig:all_scenarios} shows the switch frame for all scenarios in the Physics-IQ dataset.

Both multiframe and single-frame conditioned video models can additionally be conditioned on a human-written text description of the scene that describes the conditioning part without, however, giving away the answer of how the future unfolds. For evaluating image-to-video (i2v) and multiframe video models, we provide both the captions and the conditioning frame(s) as conditioning signals. Stable Video Diffusion is the only model in our study that does not accept text as a conditioning signal.

\oursubsection{Why create a real-world Physics-IQ dataset}
\label{subsec:related_work}
The question of whether video generative models can understand physical principles has been explored through a range of benchmarks designed to evaluate physical reasoning. Physion \cite{physion} and its successor Physion++ \cite{tung2024physionpp} use object collisions and stability to assess a model's ability to predict physical outcomes and infer relevant properties of objects (e.g., mass, friction) during dynamic interactions. Similarly, CRAFT \cite{ates2020craft} and IntPhys \cite{riochet2018intphys} assess causal reasoning and intuitive physics, testing whether models can infer forces or understand object permanence. Intuitive physics has a rich history in Cognitive Science and is concerned with understanding how humans build a commonsense intuition for physical principles \cite[e.g.][]{mccloskey1980curvilinear,mccloskey1983intuitive,kellman1983perception,spelke1992origins,spelke1995spatiotemporal,gopnik2004theory,saxe2006perception,agrawal2016learning,kubricht2017intuitive,tenenbaum2011grow,piloto2022intuitive}. Recent efforts have extended physical reasoning evaluation to generative video models. VideoPhy \cite{bansal2024videophy} and PhyGenBench \cite{meng2024world} focus on assessing physical commonsense through text-based descriptions rather than visual data. These works emphasize logical reasoning about physical principles but do not incorporate real-world videos or dynamic visual contexts. PhysGame \cite{cao2024physgame} focuses on gameplay, while the Cosmos project \cite{agarwal2025cosmos} aims to enable better embodied AI, including robotics. LLMPhy \cite{cherian2024llmphy} combines a large language model with a non-differentiable physics simulator to iteratively estimate physical hyperparameters (e.g., friction, damping, layout) and predict scene dynamics. Other benchmarks, such as CoPhy \cite{baradel2019cophy} and CLEVERER \cite{yi2019clevrer}, emphasize counterfactual reasoning and causal inference in video-based scenarios. ESPRIT \cite{rajani2020esprit} couples physical reasoning tasks with explainability via natural language explanations, and PhyWorld \cite{phyworld} evaluates the ability of generative video models to encode physical laws, focusing on physical realism. A comprehensive overview of recent models and methods is provided by \cite{liu2025generative}.

\begin{figure}[ht]
    \centering
    \includegraphics[width=\columnwidth]{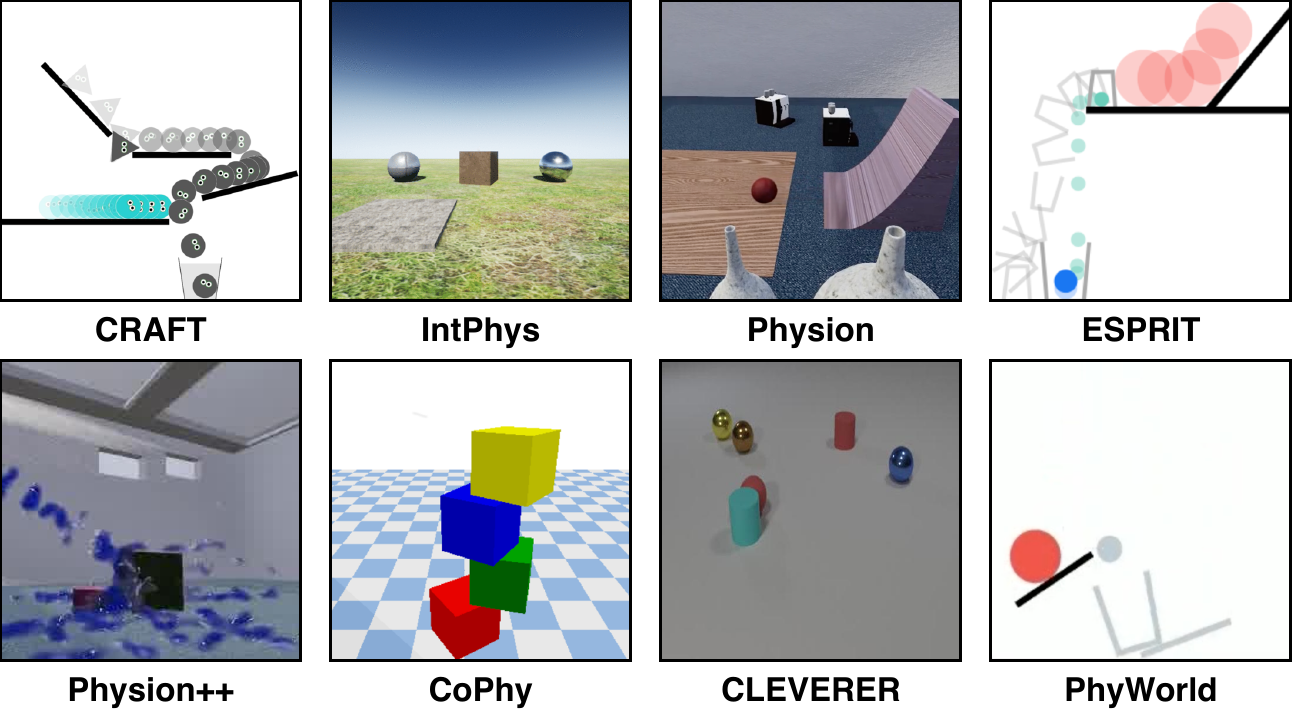}
    \caption{A qualitative overview of recent synthetic datasets related to physical understanding \cite{ates2020craft,riochet2018intphys,physion,rajani2020esprit,tung2024physionpp,baradel2019cophy,yi2019clevrer,phyworld}. These datasets are great for the purposes they were designed for, but not ideal for evaluating models trained on real-world videos due to the distribution shift.}
    \label{fig:other_datasets}
\end{figure}

However, a major drawback of many benchmarks is that the data they use is synthetic (see Fig~\ref{fig:other_datasets} for samples). This introduces a real-vs-synthetic distribution shift that may confound results when testing video models trained on natural videos. The Physics-IQ dataset overcomes this limitation by providing real-world videos, capturing diverse and complex physical phenomena (see Fig~\ref{fig:our_dataset_qualitative}). With three views per scenario, controlled and measured physical variance (by recording two takes for each video), and challenging out-of-distribution settings it provides a  rigorous design for evaluating video generative models.

\oursubsection{Models}
\label{subsec:models}
We evaluate eight different video generative models on our benchmark: VideoPoet (both i2v and multiframe) \citep{videopoet}, Lumiere (i2v and multiframe) \citep{bartal2024lumiere}, Runway Gen 3 (i2v) \citep{runway2024}, Pika 1.0 (i2v) \citep{pikalabs2024}, Stable Video Diffusion (i2v) \citep{blattmann2023stable}, and Sora (i2v) \citep{sora_openai}. We note that Luma \citep{lumaai2024} and Veo2 \citep{veo2_deepmind} are two other popular video generative models that have not been included in our benchmark, because the Luma Labs usage policy prohibits benchmarking and because Veo2 is not generally available at the time of writing. Different models have different requirements for the input conditions (single frame, multi frame, or text conditioning), frame rates (8--30 FPS), and resolution (between $256 \times 256$ and $1280 \times 768$). An overview is shown in \Cref{tab:model_overview}. For our study, we matched the model's preferred input conditions, frame rates, and resolution exactly by performing a pre-processing step on the Physics-IQ videos (see \Cref{supp:change_fps_pseudocode} for pseudocode).

VideoPoet and Lumiere are the only two models in our study that can take multiple frames as conditioning input. These models also include a super-resolution stage, where they first generate a lower resolution video and subsequently upscale it to a higher resolution. Since we noticed that the lower resolution outputs suffice to test physical realism, we skipped the super-resolution step for these models. The benchmark consists of physical interactions where temporal information is decidedly useful to have, thus it is generally to be expected that multiframe models should, in principle, be able do better than i2v models.

\oursubsection{Metrics for physical understanding}
\label{subsec:metrics}
Video generative models commonly use metrics \cite{hore2010image, wang2004image, fvd, zhang2018unreasonable} and benchmarks \cite{huang2024vbench, huang2024vbench++, he2024videoscore} suited for evaluating the visual quality and realism of the generated videos. These metrics include Peak Signal-to-Noise Ratio (PSNR) \citep{hore2010image}, Structural Similarity Index Measure (SSIM) \citep{wang2004image}, Fréchet Video Distance (FVD) \citep{fvd, Ge_2024_CVPR}, and Learned Perceptual Image Patch Similarity (LPIPS) \citep{zhang2018unreasonable}. These metrics are useful for comparing the appearance, temporal smoothness, and statistics of generated videos with the ground truth. Unfortunately, these metrics are not equipped to assess the understanding of physical laws by video models.   For instance, both PSNR and SSIM evaluate pixel-level similarities but are not sensitive to the correctness of motion and interactions in a video; FVD captures overall feature distributions but does not penalize a model for physically implausible actions and LPIPS focuses on human-like perception of similarity rather than physical plausibility. While these metrics are great for measuring what they were designed for, they are not equipped to judge whether a model understands real-world physics.

In our benchmark, we use the following four metrics to track different aspects of physical understanding:
\begin{itemize}[noitemsep]
    \item \emph{Where} does action happen? \textbf{Spatial IoU}
    \item \emph{Where} \& \emph{when} does action happen? \textbf{Spatiotemporal IoU}
    \item \emph{Where} \& \emph{how much} action happens? \textbf{Weighted spatial IoU}
    \item \emph{How} does action happen? \textbf{MSE}
\end{itemize}
These four metrics---explained in detail below---are then combined into a single score, the \textbf{Physics-IQ score}, by summing the individual scores (with a negative sign for MSE where lower= better). This Physics-IQ score is normalized such that physical variance---the upper limit of what we can reasonably expect a model to capture---is at 100\%.

\begin{figure*}[ht]
    \centering
    \includegraphics[width=\textwidth]{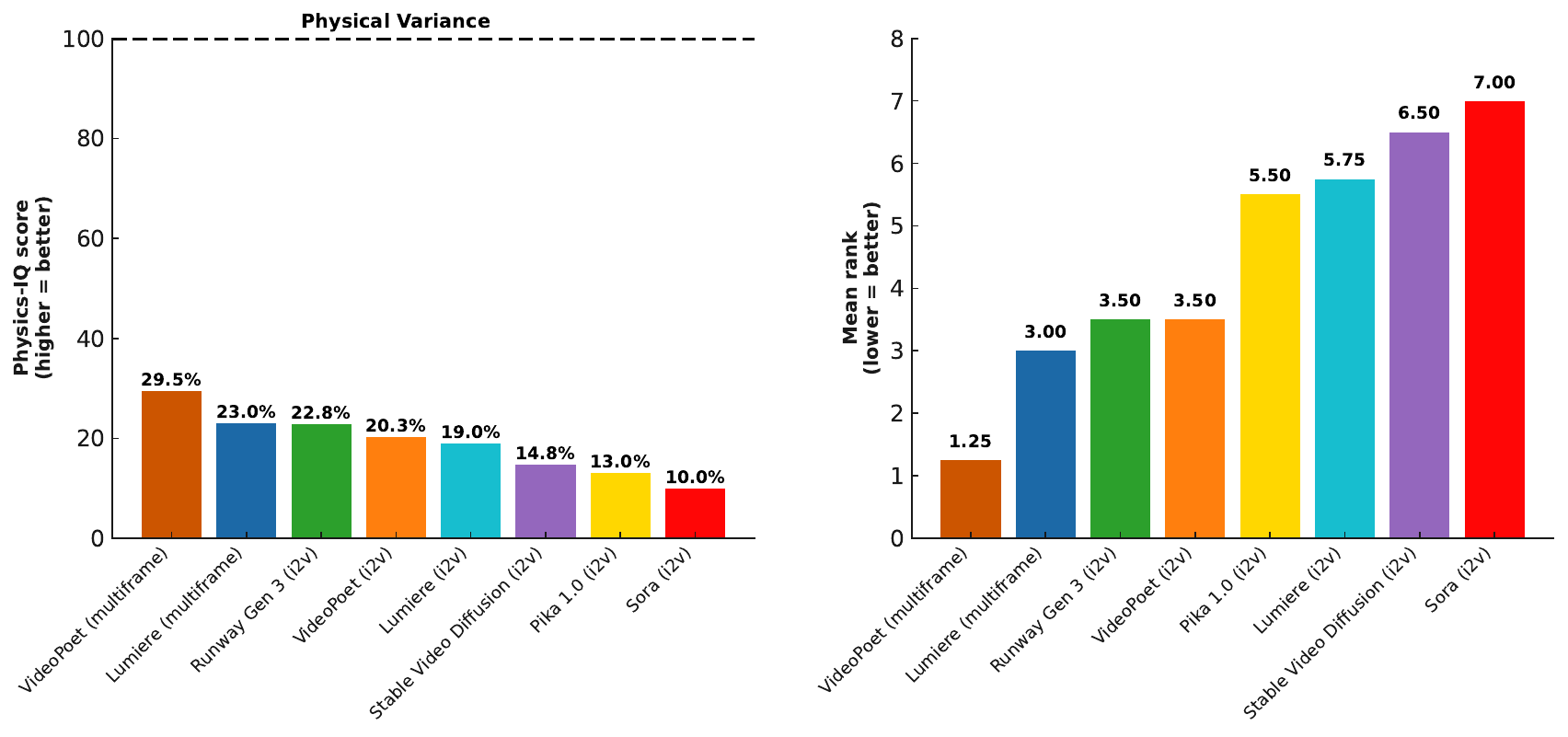}
    \caption{How well do current video generative models understand physical principles? \textbf{Left.}~The Physics-IQ score is an aggregated measure across four individual metrics, normalized such that pairs of real videos that differ only by physical randomness score 100\%. All evaluated models show a large gap, with the best model scoring 29.5\%, indicating that physical understanding is severely limited. \textbf{Right.}~In addition, the mean rank of models across all four metrics is shown here; the Spearman correlation between aggregated results on the left and mean rank on the right is high ($\text{-}.92, \emph{p}<.005$), thus aggregating to a single Physics-IQ score largely preserves model rankings.}
    \label{fig:model ranking}
\end{figure*}

\paragraph{\emph{Where} does action happen? Spatial IoU}
The location of movement is an important indicator of physical ``correctness''. For instance, in the ``domino with duck interrupting the chain'' scenario from \Cref{fig:our_dataset_qualitative}, only the part of the chain that is to the right side of the duck should tumble, while the other part should remain unchanged. Similarly, the spatial trajectory of a moving ball is indicative of whether the movement is realistic. Our Spatial IoU metric compares generated videos against ground truth to determine whether the location of movement/action mirrors ground truth. Since the benchmark videos are filmed from a static perspective without camera movement, a simple threshold on pixel intensity changes across frames (see \Cref{supp:binary_mask} for pseudocode) easily identifies where movement happens. This leads to a binary $h \times w \times t$ ``motion mask video'' that highlights the regions of motion in a scene at any point in time. Spatial IoU then simply generates a binary $h \times w$ spatial ``motion map''---similar, in spirit, to a saliency map---by collapsing the masks across the time dimension with a max operation. A motion map thus simply has a 1 whenever action occurred, at any point in time, at a particular location; and a 0 otherwise. This motion map is compared against the motion map from the real, ground truth video, using Intersection over Union or IoU (a metric commonly used in object detection to measure overlap while penalizing areas that differ):

\[
\mathsf{Spatial}\text{-}\mathsf{IoU} = \frac{|M_{\mathsf{real}}^{\mathsf{binary,spatial}} \cap M_{\mathsf{gen}}^{\mathsf{binary,spatial}}|}{|M_{\mathsf{real}}^{\mathsf{binary,spatial}} \cup M_{\mathsf{gen}}^{\mathsf{binary,spatial}}|}
\]
where \( M_{\mathsf{real}}^{\mathsf{spatial}} \) and \( M_{\mathsf{gen}}^{\mathsf{spatial}} \) are the motion maps based on real and generated videos, respectively. Spatial IoU measures whether the location \emph{where} an action happens is correct.

\paragraph{\emph{Where} \& \emph{when} does action happen? Spatiotemporal IoU}
Spatiotemporal IoU goes a step further than Spatial IoU by also taking into account \emph{when} an action occurs. Instead of collapsing across time as Spatial IoU does, Spatiotemporal IoU compares the two motion mask videos (based on real and generated videos) frame-by-frame, averaging across $t$:
\[
\mathsf{Spatiotemporal}\text{-}\mathsf{IoU}(M_{\mathsf{real}}, M_{\mathsf{gen}}) = \frac{|M_{\mathsf{real}} \cap M_{\mathsf{gen}}|}{|M_{\mathsf{real}} \cup M_{\mathsf{gen}}|}
\]
where \( M_{\mathsf{real}} \) and \( M_{\mathsf{gen}} \) are the $h \times w \times t$ binary motion masks for the real and generated videos, respectively. Spatiotemporal IoU thus tracks not only \emph{where} an action occurs in a video, but also whether it occurs at the right time (\emph{when}). If a model does well on Spatial IoU but poorly on Spatiotemporal IoU, this would therefore indicate that the model gets the location of the action right, but the timing wrong.

\paragraph{\emph{Where} does action happen, and \& \emph{how much} action happens? Weighted spatial IoU}
Weighted spatial IoU is similar to Spatial IoU in the sense that it compares two $h \times w$ ``motion maps''. However, instead of comparing binary motion maps (action occurred or did not occur), it also assesses \emph{how much} action happens at any given location. This distinguishes between e.g.\ motion caused by a pendulum (showing repeated motion in an area) from motion by a rolling ball (which passes a location only once). Weighted spatial IoU is computed by taking the binary $h \times w \times t$ motion mask video (described above in the section on Spatial IoU) and collapsing across the time dimension $t$ in a weighted fashion (instead of taking the maximum). The weighting simply averages per-frame action. This weighted $h \times w$ spatial ``motion map'' is then used to compute the metric by summing the pixel-wise minimum of two motion maps and dividing by the pixel-wise maximum:
\[
\mathsf{Weighted}\text{-}\mathsf{spatial}\text{-}\mathsf{IoU} = \frac{\sum_{i=1}^n \min\big(M_{\mathsf{real, i}}^{\mathsf{weighted,spatial}}, M_{\mathsf{gen, i}}^{\mathsf{weighted,spatial}}\big)}{\sum_{i=1}^n \max\big(M_{\mathsf{real, i}}^{\mathsf{weighted,spatial}}, M_{\mathsf{gen, i}}^{\mathsf{weighted,spatial}}\big)}
\]
where \( M_{\mathsf{real}}^{\mathsf{weighted,spatial}} \) and \( M_{\mathsf{gen}}^{\mathsf{weighted,spatial}} \) are the weighted motion maps representing how much activity/action happes at any location (based on real and generated videos, respectively).
Weighted spatial IoU thus measures not only \emph{where} an action occurs, but also \emph{how much} action is happening.

\paragraph{\emph{How} does an action happen? MSE}
Finally, mean squared error (MSE) calculates the average squared difference between corresponding pixel values in two frames (e.g., a real and a generated frame). Given two frames \( f_{\mathsf{real}} \) and \( f_{\mathsf{gen}} \), the MSE is given by:
\begin{align*}
    \mathsf{MSE}(f_{\mathsf{real}}, f_{\mathsf{gen}}) = \frac{1}{n} \sum_{i=1}^{n} (f_{\mathsf{real, i}} - f_{\mathsf{gen, i}})^2
\end{align*}
where \( n \) is the total number of pixels in the frame. MSE focuses on pixel-level fidelity; this is a very strict requirement that is sensitive to \emph{how} objects look and interact. For instance, if a generative model would show a tendency to change the color of objects, this physically unrealistic event would be heavily penalized. MSE therefore tracks aspects that complement the three other metrics. None of them is perfect, and none of them should be used in isolation, but collectively they provide a comprehensive assessment of different aspects that quantify physical realism. Since raw MSE values can be hard to interpret, we provide an intuition in \Cref{fig:mse_visualization}.

\begin{figure*}[t]
  \centering
  \begin{subfigure}{0.48\textwidth}
    \includegraphics[width=\textwidth]{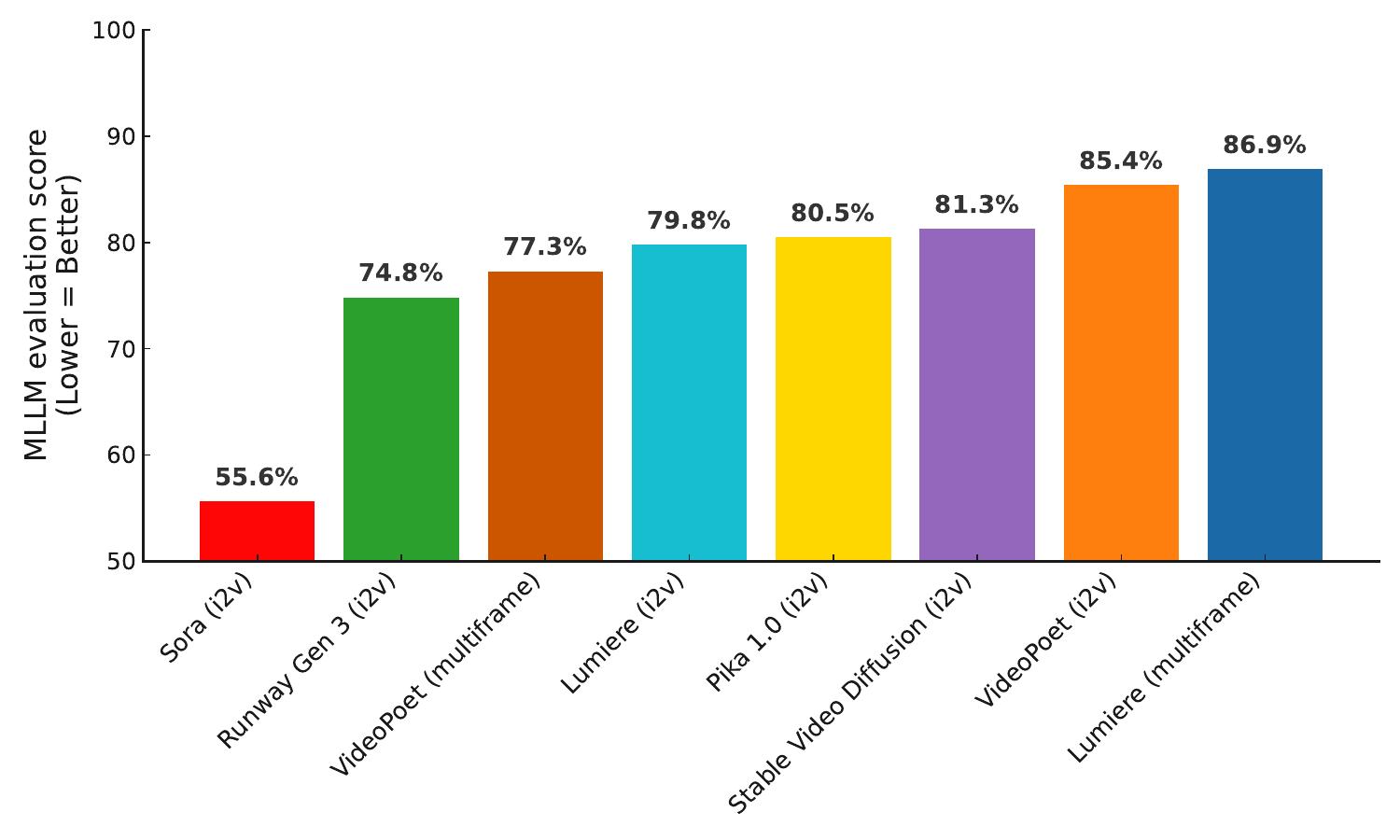}
    \label{subfig:mllm_barplot}
  \end{subfigure}
  \hfill
  \begin{subfigure}{0.48\textwidth}
    \includegraphics[width=\textwidth]{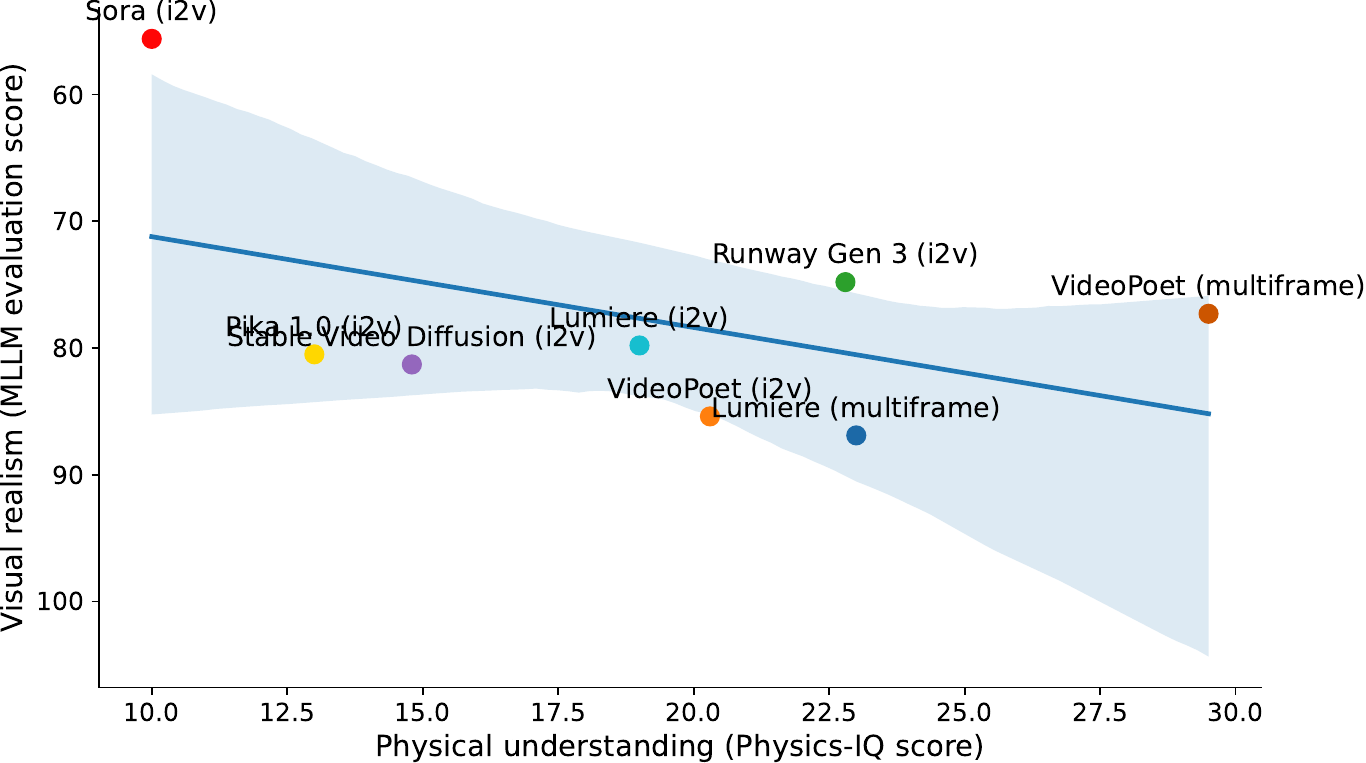}
    \label{subfig:correlation_plot}
  \end{subfigure}
  \caption{Relationship between visual realism and physical understanding. \textbf{Left.} A multimodal large language model (Gemini 1.5 Pro) is asked to identify the generated video among the real and the generated video for each scenario (MLLM score) in a two-alternative forced choice paradigm. Chance rate is 50\%; lower scores indicate that the model finds it harder to tell apart generated from real videos (= better realism). Sora-generated videos are hardest to distinguish from real videos for the model, whereas Lumiere (multiframe) is easiest. \textbf{Right.} Do models that produce `realistic-looking' videos (as assessed by the MLLM score) also score better in terms of physical understanding (as assessed via the Physics-IQ score)? This scatterplot with linear fit and 95\% confidence interval as a shaded blue area shows that this is not the case: Visual realism is uncorrelated with physical understanding (Pearson's \textit{r} = - 0.46, \textit{p}=.249 not significant). Note that the y axis on this plot is inverted for easier interpretation (up \& right are best).}
  \label{fig:realism_vs_understanding}
\end{figure*}

\oursubsection{Metric for visual realism: MLLM evaluation}
In addition to measuring the physical realism, we are interested in tracking how convincingly a model can generate realistic videos, as assessed by a multimodal large language model or MLLM (in our case: Gemini 1.5 Pro, \cite{gemini2024}). Instead of rating videos (which would be sensitive to model biases), we use the gold standard experimental methods from psychophysics, a 2AFC paradigm. 2AFC stands for two-alternative-forced-choice. In our case, this means that the MLLM is given pairs of real and generated videos of the same scenario in randomized order. The MLLM is asked to identify the generated video. The MLLM evaluation score is expressed as a percentage corresponding to the accuracy across all videos, with chance rate at 50\%. Any accuracy that is higher indicates that the MLLM was able to correctly identify at least some of the generated videos; while accuracies close to 50\% indicates that a video generative model has successfully deceived the MLLM into classifying the generated videos as real, indicating high visual realism. Details on the experiment are described in the appendix.

\oursection{Results}

\begin{figure*}[ht!]
    \centering
    \includegraphics[width=\textwidth]{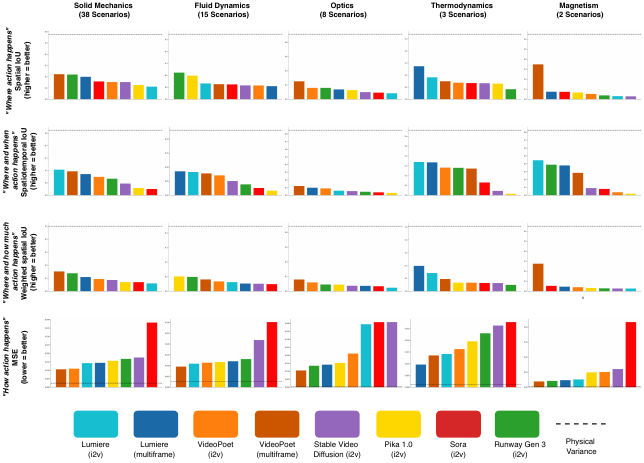}
    \caption{Performance comparison of video generative models across different physical categories (columns) and metrics (rows). For the top three metrics, higher is better; for the last metric lower values are best. Throughout, physical variance (i.e., the performance that is achievable by real videos differing only by physical randomness) is indicated by a dashed line. Across metrics and categories, models show a considerable lack in physical understanding. More lenient metrics like $\mathsf{Spatial}\text{-}\mathsf{IoU}$ (top row) that only assess \emph{where} an action occurred lead to higher scores than more strict metrics that also take into account e.g.\ \emph{when} or \emph{how much} action should be taking place.
}
    \label{fig:category_breakdown}
\end{figure*}

\begin{figure*}[ht!]
    \centering
    \includegraphics[width=\textwidth]{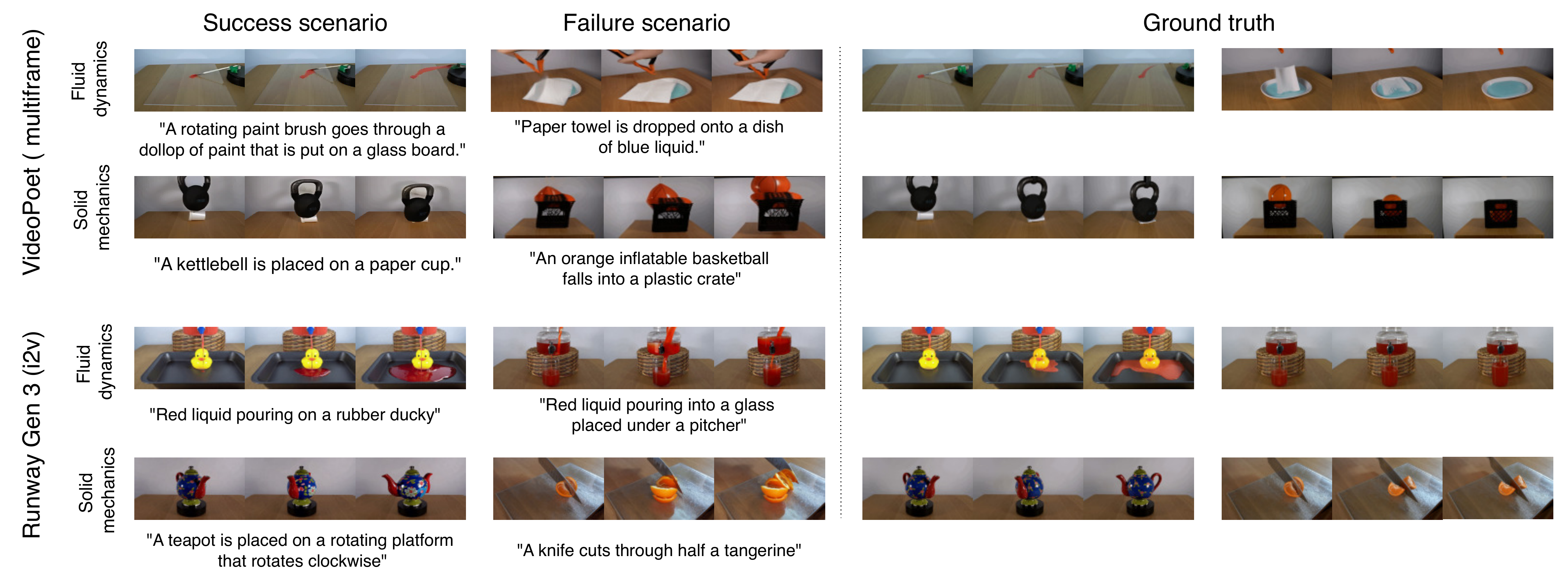}
    \caption{We here visualize success and failure scenarios within the fluid dynamics and solid mechanics categories for the two best models, VideoPoet and Runway Gen 3, according to our metrics. Both models are able to generate physics plausible frames for scenarios such as smearing paint on glass (VideoPoet) and pouring red liquid on a rubber duck (Runway Gen 3). At the same time, the models fail to simulate a ball falling into a crate or cutting a tangerine with a knife. See \extlink{https://physics-iq.github.io/figures/fig2.html}{here} for an animated version of this figure.}
    \label{fig:best_worst_visualize}
\end{figure*}
\oursubsection{Physical understanding}
The goal of our Physics-IQ benchmark is to understand, and quantify, whether generative video models learn physical principles. Therefore, we test all eight models in our study on every scenario and for each camera position (left, center, right) in the benchmark dataset. These samples are visualized in \Cref{fig:our_dataset_qualitative}. We first report the aggregated Physics-IQ results across all metrics related to physical understanding ($\mathsf{Spatial}\text{-}\mathsf{IoU}, ~\mathsf{Spatiotemporal}\text{-}\mathsf{IoU}, ~\mathsf{Weighted}\text{-}\mathsf{spatial}\text{-}\mathsf{IoU}, ~\mathsf{MSE}$) in \Cref{fig:model ranking}. The main takeaway from the left part of this figure is that all the models show a massive gap to the physical variance baseline, with the best model scoring only 29.5\% out of the possible 100.0\%. As we mentioned in the previous section, each scenario was recorded twice (\textit{take 1} and \textit{take 2}) to estimate the natural variability in real-world physical phenomena. This estimate is termed the \emph{physical variance}; the figure is normalized such that pairs of real videos that differ only by physical randomness score 100.0\%. The gap between model performance and real videos demonstrates a severe lack of physical understanding in current powerful video generative models. Across the different models, VideoPoet~(multiframe) \cite{videopoet} ranks best; interestingly, VideoPoet is a causal model. For the two models that have both an image2video (i2v) and a version conditioned on multiple frames (multiframe), the multiframe variants do better than the i2v variants. This is to be expected given that in order to predict the future as we require models to do on our challenging Physics-IQ benchmark, having access to temporal information (as multiframe variants have) should help.
Zooming in on these overall results, \Cref{fig:category_breakdown} breaks down model performance into different categories that include solid mechanics, fluid dynamics, optics, thermodynamics, and magnetism. While there is no category that can be considered ``solved'', performance varies across categories, with some showing promising indications and differences across models. Interestingly, all models perform much better on $\mathsf{Spatial}\text{-}\mathsf{IoU}$, a metric that has the weakest requirement in the sense that it is only sensitive to \emph{where} an action occurred, not whether it occurred at the right time (as $\mathsf{Spatiotemporal}\text{-}\mathsf{IoU}$ would track) or whether it had just the right \emph{amount} of action (as measured by $\mathsf{Weighted}\text{-}\mathsf{spatial}\text{-}\mathsf{IoU}$). Furthermore, even a relatively simple metric like $\mathsf{MSE}$ shows a large gap between physically realistic videos and model-generated predictions.
The performance of each model on each individual metric is reported in \Cref{tab:model_metrics}. As expected based on the aggregated results, VideoPoet (multiframe) performs best on a majority of the metrics (3 out of 4). Qualitatively, the generated videos from Sora are often visually and artistically superior, but they also frequently include transition cuts---despite instructed not to change the camera perspective---which is penalized by several other metrics. We expect that if a future version of this model more closely follows the prompt (static camera perspective, no camera movement), its Physics-IQ score would improve substantially. Qualitatively, success and failure cases are visualized in \Cref{fig:best_worst_visualize}.

\begin{table}[h]
\centering
\caption{Comparison of metric scores for different models. The best-performing model for each metric is marked in bold. Note that Physical Variance serves as a performance upper bound for each metric, representing the difference between two real videos and capturing the inherent variability in real-world scenarios.} %
\resizebox{1.0\columnwidth}{!}{
\begin{tabular}{|l|c|c|c|c|c|}
\hline
\textbf{Model} & \makecell{\textbf{Spatial} \\ \textbf{IoU} $\uparrow$} & \makecell{\textbf{Spatiotemporal} \\ \textbf{IoU} $\uparrow$} & \makecell{\textbf{Weighted spatial} \\ \textbf{IoU} $\uparrow$} & \makecell{\textbf{MSE} \\ $\downarrow$} & \makecell{\textbf{Physics-IQ Score} \\ $\uparrow$} \\
\hline
\cellcolor{gray!20}\textbf{Physical Variance} & \cellcolor{gray!20}\textbf{0.678} & \cellcolor{gray!20}\textbf{0.535} & \cellcolor{gray!20}\textbf{0.577} & \cellcolor{gray!20}\textbf{0.002} & \cellcolor{gray!20}\textbf{100.0} \\
\hline
\cellcolor[HTML]{CC5500}\textbf{VideoPoet (multiframe)} & \textbf{0.204} & 0.164 & \textbf{0.137} & \textbf{0.010} & \textbf{29.5} \\
\hline
\cellcolor[HTML]{1C69A7}\textbf{Lumiere (multiframe)} & 0.170& 0.155 & 0.093 & 0.013 & 23.0 \\
\hline
\cellcolor[HTML]{2ca02c}\textbf{Runway Gen 3 (i2v)} & 0.201 & 0.115 & 0.116 & 0.015 & 22.8 \\
\hline
\cellcolor[HTML]{FF7F0E}\textbf{VideoPoet (i2v)} & 0.141 & 0.126 & 0.087 & 0.012 & 20.3 \\
\hline
\cellcolor[HTML]{17BECF}\textbf{Lumiere (i2v)} & 0.113 & \textbf{0.173} & 0.061 & 0.016 & 19.0 \\
\hline
\cellcolor[HTML]{9467BD}\textbf{Stable Video Diffusion (i2v)} & 0.132 & 0.076 & 0.073 & 0.021 & 14.8 \\
\hline
\cellcolor[HTML]{FFD700}\textbf{Pika 1.0 (i2v)} & 0.140 & 0.041 & 0.078 & 0.014 & 13.0 \\
\hline
\cellcolor[HTML]{ff0606}\textbf{Sora (i2v)} & 0.138 & 0.047 & 0.063 & 0.030 & 10.0 \\
\hline
\end{tabular}
}
\vspace{0.25cm}
\label{tab:model_metrics}
\end{table}

\oursubsection{Visual realism: Multimodal large language model evaluation}
Our results indicate a striking lack of physical understanding in current generative video models. Why, then, do samples from many of those models that are circulated online look so realistic? We decided to quantify the \emph{visual realism} of model-generated videos by asking a capable multimodal large language model, Gemini 1.5 Pro \cite{gemini2024}, to identify the generated one out of a pair of two videos for each scenario in the Physics-IQ benchmark). The result of this experiment is presented in \Cref{fig:realism_vs_understanding} (left). The MLLM score evaluates a model's ability to generate videos that can deceive a multimodal large language model into classifying them as real. Accuracies that are closer to chance performance (50\% by randomly guessing) indicate that the MLLM finds it harder to tell apart real from generated videos, thus attesting to the visual realism of the generated video. Overall, the MLLM frequently succeeds in identifying the model-generated video (up to 86.9\% accuracy for Lumiere multiframe); that said, the model-produced explanations for decisions are often unrelated to the visual content, akin to post-hoc rationalizations known from human experiments \cite{nisbett1977telling}. One model particularly stands out: Sora achieved the best MLLM score of 55.6\%, outperforming all other models by a significant margin. Runway Gen 3 and VideoPoet (multiframe) ranked second and third, with scores of 74.8\% and 77.3\%, respectively, albeit with a considerable gap behind Sora. Thus, highly capable models such as Sora indeed succeed in generating visually realistic videos---even though their ability to understand physical principles, as shown in the previous section, is very limited. This finding aligns with a number of intriguing studies discovering that many deep learning models also lack an understanding of intuitive physics and causal reasoning \cite{lake2017building,storks2021tiered,weihs2022benchmarking,binz2023using,jassim2023grasp,schulze2025visual} and are sometimes described as ``blind'' \cite{rahmanzadehgervi2024vision}. For video models, we find no significant correlation between \emph{visual realism} and \emph{physical understanding}: \Cref{fig:realism_vs_understanding} (right) demonstrates that there is a distinction between generating realistic videos and comprehending the physical principles of reality.
\oursection{Discussion}
We introduced Physics-IQ, a challenging and comprehensive real-world benchmark to evaluate physical understanding in video generative models. We analyzed eight models on Physics-IQ and proposed metrics to quantify physical understanding. The benchmark data and metrics cover a wide range of settings and reveal a striking discrepancy between visual realism (sometimes present in current models) and physical understanding (largely lacking in current models).

\paragraph{Do video models understand physical principles?}
We investigated the question whether the ability of video generative models to generate realistic-looking videos also implies that they have acquired an understanding of the physical principles that govern reality. Our benchmark shows that this is not the case: all evaluated models currently lack a deep understanding of physics. Even the highest-scoring model, VideoPoet (multiframe), only achieves a score of 29.5, falling significantly short of the best possible score of 100.0 obtained from the \emph{physical variance} baseline, which quantifies the natural variability observed between real-world videos. That said, our results don't imply that future models cannot learn a better physical understanding through next frame prediction. It remains an open question whether simply scaling the same paradigm further might solve this, or whether alternative (and possibly more interactive) training schemes might be required. Given the success of scaling deep learning, we are optimistic that future-frame prediction alone could lead to a much better understanding: while successful prediction does not guarantee successful understanding, acquiring a better understanding should certainly help with successful prediction. Learning physics by predicting pixels may sound challenging, but language models are known to learn syntax and grammar from text prediction alone \cite{hewitt2019structural}.

It may be worth pointing out that even though the models in our study often failed to generate physically plausible continuations, most current models were already successful on some scenarios. For example, the highest-ranking model, VideoPoet (multiframe), displayed remarkable physical understanding in certain scenarios---such as accurately simulating paint smearing on glass. In contrast, many lower-ranking models exhibited fundamental errors, such as physically implausible interactions (e.g., objects passing through other objects). A studey based on synthetic datasets \citep{phyworld} has shown that given a large enough dataset size, video models are able to learn specific physical laws. We consider it likely that as models are trained on larger and more diverse corpora of videos, their understanding of real-world physics will continue to improve. We hope that quantifying physical understanding, as we aim to do by open-sourcing the physics-IQ benchmark, might facilitate tracking progress in this area.

\paragraph{Visual realism doesn't imply physical understanding.} 
We observed a striking discrepancy between visual realism and physical understanding: those two properties are not statistically significantly correlated (\Cref{fig:realism_vs_understanding}), thus models that produce the most realistic-looking videos don't necessarily show the most physically-plausible continuations. For instance, in a scenario where a burning matchstick is lowered into a glass full of water (leading to the flame being extinguished), Runway Gen 3 generates a continuation where as soon as the flame touches the water, a candle spontaneously appears and is lit by the match. Every single frame of the video is high quality in terms of resolution and realism, but the temporal sequence is physically impossible. Such a tendency to hallucinate objects into existence is a drawback of many generative models and an active area of research in deep learning \citep{rawte2023survey}. In our experiments, we observed hallucinations in all models, but more powerful models like Runway Gen 3 and Sora often hallucinated information that was at least consistent with the scenario (e.g.\ matchstick - candle), indicating at least a certain level of understanding.

\paragraph{Dataset biases are reflected in model generations.}
We observed that most models were consistent in producing videos that aligned with the scene and viewpoint provided. Models like Sora and Runway Gen 3 were particularly strong at understanding the given scene and generating subsequent frames that were consistent with respect to object placement and their attributes (shape, color, dynamics). Interestingly, many models also demonstrated biased generations depending on properties of their training. For example, in prototyping experiments we observed that when given a conditioning video of a red pool table where one ball hits another, as Lumiere starts generating, it immediately turned the red pool table to a green one, showing a bias to commonly occurring green pool tables. Similarly, videos generated by Sora often featured transition cuts, possibly suggesting a training paradigm optimized to generate artistic videos.

\paragraph{Metrics and their limitations.} Popular metrics to test quality and measure realism of generated videos include PSNR \citep{hore2010image}, FVD \citep{fvd}, LPIPS \citep{zhang2018unreasonable} and SSIM \citep{wang2004image}. However, designing metrics that quantify physical understanding in generated videos is a challenging endeavor. We proposed a suite of metrics to measure the spatial, temporal and perceptual coherence of video models. While individually none of these metrics is perfect, the combined insights from these metrics and the \emph{Physics-IQ score} that integrates a normalized score across these metrics provide a holistic assessment of the strengths and weaknesses of video models. That said, none of these metrics directly quantify any physical phenomena and instead serve as proxies. For instance, the MLLM metric provided a way to quantify how much generated videos `deceive' a multimodal model. However, the metric is limited by the capability of the underlying multimodal large language model (MLLM) itself. In our analysis, we found that while the MLLM was frequently able to identify generated videos (except for videos generated by Sora), its explanations for the decision were often wrong. As another example, we observed that Stable Video Diffusion often generated videos with large amounts of hallucinations and implausible object motions; nonetheless, its $\mathsf{Spatial}\text{-}\mathsf{IoU}$ score is in the same ballpark as Lumiere, Sora, Pika and VideoPoet (i2v) showing that no metric should be assessed in isolation. This is also evidenced by the fact that e.g.\ Runway Gen 3 did very well in terms of the spatial location of actions ($\mathsf{Spatial}\text{-}\mathsf{IoU}$) while scoring poorly on temporal consistency ($\mathsf{Spatiotemporal}\text{-}\mathsf{IoU}$).

We intentionally designed Physics-IQ to be a challenging benchmark in order to provide useful signal for model development in the future; in this context it may be worth noting that our metrics may be on the conservative side by strongly penalizing object hallucinations, camera movement (which we prompted models not to do) or shot changes. For instance, Sora tends to show these more frequently than other models, leading to low scores on some metrics. This is not ideal, but we believe that in an area like deep learning where hype is omnipresent, scientific benchmarks should err on the side of caution.

\paragraph{Outlook: Understanding without interaction?} Our findings are connected to a larger, interdisciplinary debate at the heart of intelligence: Does an understanding of the world emerge from predicting what happens next (next-video-frame prediction in artificial intelligence, predictive coding in neuroscience)---or, alternatively, is it necessary to interact with the world in order to understand it (as argued by proponents of embodied cognition and robotics)? In cognitive science, being able to interact with the world is seen as an important component for developing intuitive physics \cite{spelke1988origins,baillargeon2002acquisition,vicovaro2023grounding, chang2016compositional}, in combination with predicting the outcome of a person's actions \cite{fischer2021tool, li2023approximate, sosa2025blending}. In contrast, deep learning's current bread-and-butter approach is scaling models and datasets without interactions. Will these models essentially solve physical understanding---or instead, hit a limit after which one can only improve one's understanding of the world by interacting with it? The jury is still out on this question, but the benchmark and analyses introduced in this article might help quantifying this either way. In addition to future models, improvements could also come from inference-time scaling \cite{jaech2024openai,snell2024scaling,ma2025inference} such as sampling more. If this would indeed lead to strong results, it would raise the following question: from a model's perspective, is reality but one possibility among infinitely many others?

\acknow{
The authors would like to thank David Fleet, Been Kim, Pieter-Jan Kindermans, Kelsey Allen, Jasmine Karimi, Katherine Hermann, Mike Mozer, Phoebe Kirk, Saurabh Saxena, Daniel Watson, Meera Hahn, Sara Mahdavi, Tim Brooks, Charles Herrmann, Isabelle Simpson, Jon Shlens, and Chris Jones for helpful discussions and/or supporting this project in various ways.
}

\showacknow{}

\subsection*{References}
\bibliography{refs}
\clearpage

\section*{Supplementary Material}
\oursubsection{Overview of all Physics-IQ scenarios}

\begin{figure*}[t]
    \centering
    \includegraphics[width=\textwidth]{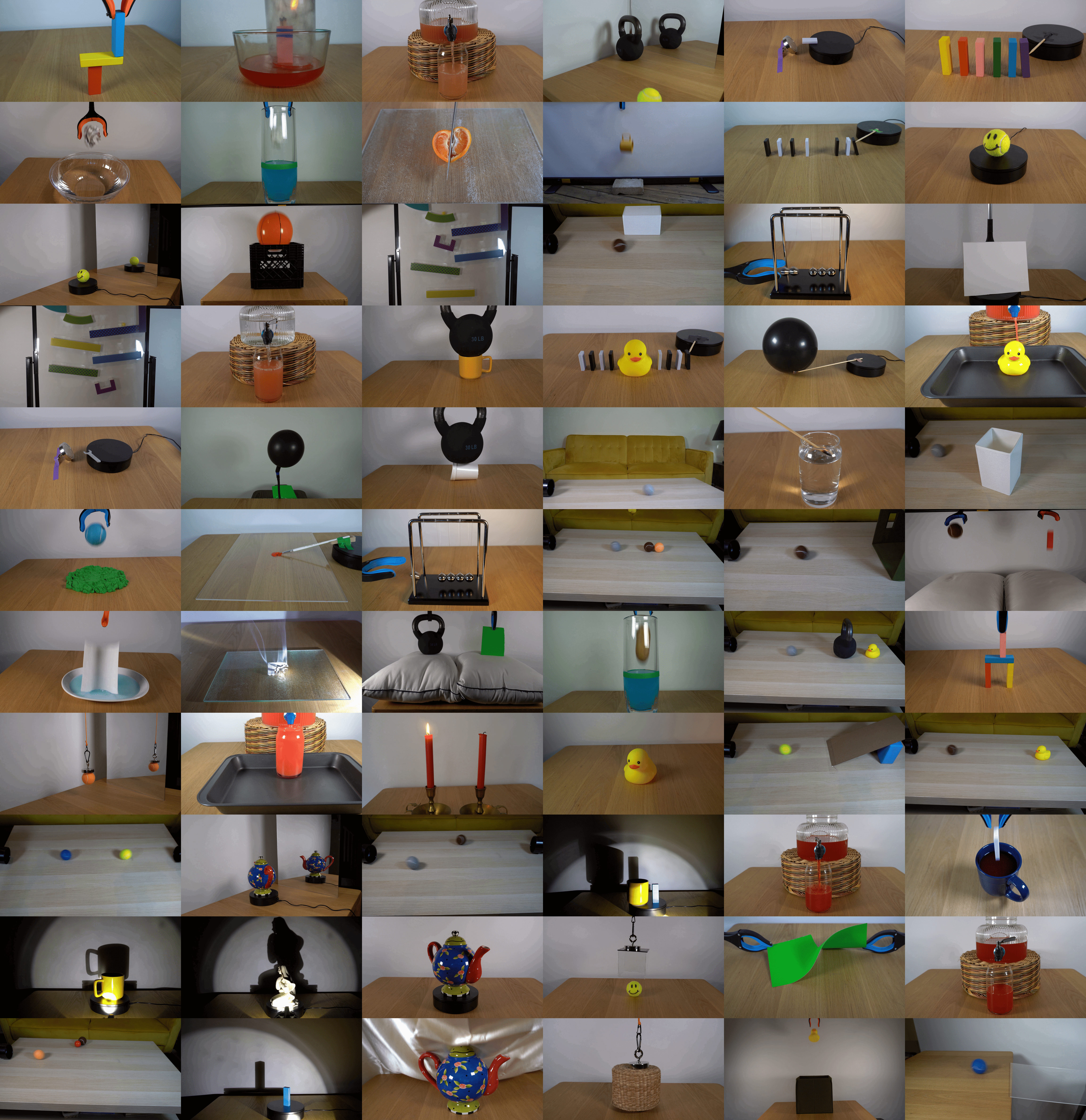}
    \caption{The switch frames (here: center view only) of all scenarios in the Physics-IQ benchmark. A switch frame is the last conditioning frame before a model is asked to predict 5 seconds of future frames.}
    \label{fig:all_scenarios}
\end{figure*}

\Cref{fig:all_scenarios} presents the switch frames (center view) from all 66 scenarios in the Physics-IQ dataset. These frames represent the last frame of the conditioning signal, after which a model is asked to generate a prediction for the future frames.

\oursubsection{Visualizing different MSE values}
\begin{figure*}[h!]
    \centering
    \includegraphics[width=\textwidth]{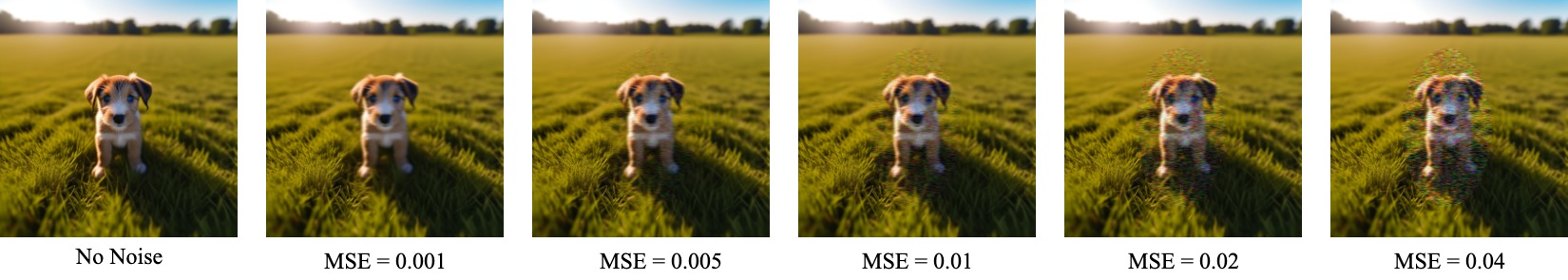}
    \caption{Since mean squared error (MSE) values can be hard to interpret, this figure shows the effect of a distortion applied to the scene, serving as a rough intuition for the effect of a MSE at different noise levels.}
    \label{fig:mse_visualization}
\end{figure*}

\Cref{fig:mse_visualization} illustrates the relationship between a distortion applied to a video and MSE (Mean Squared Error) in a scene. Note that none of the videos in the benchmark have a distortion applied to them; instead, this is inteneded as an visual intuition for how much a certain MSE value distorts an image.

\begin{table}[h]
\centering
\caption{Specifications of evaluated video models, including input conditioning, frame rate (FPS), and resolution.}
\resizebox{1.0\columnwidth}{!}{
\begin{tabular}{|l|c|c|c|c|c|}
\hline
\textbf{Model} & \makecell{\textbf{Text} \\ \textbf{Condition}} & \makecell{\textbf{Multi-frame} \\ \textbf{Condition}} & \makecell{\textbf{Single-frame} \\ \textbf{Condition}} & \textbf{FPS} & \textbf{Resolution} \\
\hline
\cellcolor[HTML]{FF7F0E}\textbf{VideoPoet (i2v)} & \checkmark & \ding{55} & \checkmark & 8 & 128×224 \\
\hline
\cellcolor[HTML]{CC5500}\textbf{VideoPoet (multiframe)} & \checkmark & \checkmark & \ding{55} & 8 & 128×224 \\
\hline
\cellcolor[HTML]{17BECF}\textbf{Lumiere (i2v)} & \checkmark & \ding{55} & \checkmark & 16 & 128×128 \\
\hline
\cellcolor[HTML]{1C69A7}\textbf{Lumiere (multiframe)} & \checkmark & \checkmark & \ding{55} & 16 & 128×128 \\
\hline
\cellcolor[HTML]{9467BD}\textbf{Stable Video Diffusion (i2v)} & \ding{55} & \ding{55} & \checkmark & 8 & 1024×576 \\
\hline
\cellcolor[HTML]{2ca02c}\textbf{Runway Gen 3 (i2v)} & \checkmark & \ding{55} & \checkmark & 24 & 1280×768 \\
\hline
\cellcolor[HTML]{FFD700}\textbf{Pika 1.0 (i2v)} & \checkmark & \ding{55} & \checkmark & 24 & 1280×720 \\
\hline
\cellcolor[HTML]{ff0606}\textbf{Sora (i2v)} & \checkmark & \ding{55} & \checkmark & 30 & 854×480 \\
\hline
\end{tabular}
}

\label{tab:model_overview}
\end{table}

\oursubsection{Adjusting video frame rate}
\begin{algorithm}[h]
\caption{Change video FPS with linear interpolation}
\begin{algorithmic}[1]
\Require video file $V$, original FPS $\mathsf{fps_{original}}$, new FPS $\mathsf{fps_{new}}$, output dimensions $(w, h)$ (optional)
\Ensure video $V'$ with adjusted FPS and resolution
\State $f_{\mathsf{original}} \gets$ extract frames from $V$ at $\mathsf{fps_{original}}$
\State $\mathsf{duration} \gets$ length of of $V$
\State $n_{\mathsf{original}} \gets$ number of frames in $\mathsf{fps_{original}}$
\State $n_{\mathsf{new}} \gets \mathsf{duration} \cdot \mathsf{fps_{new}}$
\State Initialize empty list $n_{\mathsf{new}}$
\For{$j \gets 0$ to $n_{\mathsf{new}} - 1$}
    \State $\alpha \gets j \times (n_{\mathsf{original}} - 1) / (n_{\mathsf{new}} - 1)$
    \State $i \gets \lfloor \alpha \rfloor$ \Comment{Index of the first frame for interpolation}
    \State $\beta \gets \alpha - i$ \Comment{Weight for linear interpolation}
    \State $f_1 \gets f_{\mathsf{original}}[i]$
    \State $f_2 \gets f_{\mathsf{original}}[\min(i+1, n_{\mathsf{original}} - 1)]$
    \State $f_{\mathsf{interpolated}} \gets (1 - \beta) \cdot f_1 + \beta \cdot f_2$
    \If{$(w, h)$ is not None} 
        \State resize $f_{\mathsf{interpolated}}$ to $(w, h)$
    \EndIf
    \State append $f_{\mathsf{interpolated}}$ to $f_{\mathsf{new}}$
\EndFor
\State $V' \gets$ recreate video from $f_{\mathsf{new}}$ with $\mathsf{fps_{new}}$
\State Save $V'$
\end{algorithmic}
\label{supp:change_fps_pseudocode}
\end{algorithm}

This pseudocode outlines the method for changing the frame rate (FPS) of a video using linear interpolation. It generates a smooth transition between original frames while optionally resizing the output resolution. This technique ensures temporal consistency, making it well-suited for generating videos with desired FPS to adapt Physics-IQ for models with different FPS.

\oursubsection{Generating binary mask videos}
\begin{algorithm}[h]
\caption{Generate binary mask video for moving objects}
\label{alg:binary_mask_video}
\begin{algorithmic}[1]
\Require Video $V$, output file $V'$, threshold $\tau$, update rate $\alpha$, averaging window size $w$
\Ensure Binary mask video $V'$ highlighting moving objects

\State Initialize video reader for $V$ and writer for $V'$
\State Read first $w$ frames $\{f_1, \dots, f_w\}$ and preprocess: grayscale and blur
\State Initialize background model $\text{B} \gets \frac{1}{w} \sum_{i=1}^w f_i$ \Comment{Initial average reduces noise}

\For{each frame $f_t$ in $V$}
    \State Preprocess $f_t$: grayscale and blur
    \State Update background $\text{B} \gets (1 - \alpha) \cdot \text{B} + \alpha \cdot f_t$
    \State Compute difference $d_t \gets |f_t - \text{B}|$
    \State Threshold $m_t \gets 255$ if $d_t > \tau$, else $0$
    \State Morphologically clean $m_t$ (opening and closing)
    \State Write $m_t$ to $V'$
\EndFor

\State Save and close $V'$
\end{algorithmic}
\label{supp:binary_mask}
\end{algorithm}

This pseudocode describes a method to generate binary mask videos that highlight moving objects. The algorithm combines background subtraction with adaptive updates and morphological operations to detect and cleanly segment motion in video frames. This approach is useful for creating spatial and temporal masks in Physics-IQ evaluations.

\oursubsection{MLLM evaluation prompt}
The following prompt was used in the two-alternative forced-choice paradigm: ``Your task is to help me sort my videos. I mixed up real videos that I shot with my camera and similar videos that I generated with a computer. I only know that exactly one of the two videos is the real one, and exactly one of the following two videos is the generated one. Please take a look at the two videos and let me know which of them is the generated one. I'll tip you \$100 if you do a great job and help me identify the generated one. First explain your reasoning, then end with the following statement: `For this reason, the first video is the generated one' or `For this reason, the second video is the generated one'.''

\end{document}